\documentclass[10pt,twocolumn,letterpaper]{article}

\usepackage{cvpr}

\definecolor{cvprblue}{rgb}{0.21,0.49,0.74}
\usepackage[pagebackref,breaklinks,colorlinks,allcolors=cvprblue]{hyperref}
\usepackage{adjustbox}
\usepackage{pifont}
\usepackage{xcolor}

\newcommand{\cmark}{\ding{51}}
\newcommand{\xmark}{\ding{55}}
\usepackage{tabularx}
\newcolumntype{C}{>{\centering\arraybackslash}X} 
\newcolumntype{R}{>{\raggedleft\arraybackslash}X}

\usepackage{xcolor}    
\usepackage[table]{xcolor}

\definecolor{myred}{RGB}{255, 102, 102}
\definecolor{myorange}{RGB}{255, 178, 102}
\definecolor{myyellow}{RGB}{255, 255, 153}
\definecolor{spc}{RGB}{119, 107, 170}
\definecolor{pct}{rgb}{0.7, 0, 0.2}
\newcommand{\spc}{\textcolor{spc}{$\mathbf{\circ}$\,}}
\newcommand{\pct}{\textcolor{pct}{$\bullet$\,}}

\definecolor{Qing}{rgb}{1,0.2,0.2}

\def\paperID{1616} 
\def\confName{CVPR}
\def\confYear{2026}

\title{You Only Erase Once: Erasing Anything without Bringing Unexpected Content}

\author{
    Yixing Zhu$^1$\quad  
    Qing Zhang$^{1,3*}$\quad  
    Wenju Xu$^2$\quad 
    Wei-Shi Zheng$^{1,3}$ \\
   $^1$School of Computer Science and Engineering, Sun Yat-sen University, China \quad
   $^2$Amazon \\
    $^3$Key Laboratory of Machine Intelligence and Advanced Computing, Ministry of Education, China
}

\begin{document}

\twocolumn[
    \maketitle
    \vspace{-3.0em}
    \begin{center}
    \centering
    \includegraphics[width=1.0\linewidth]{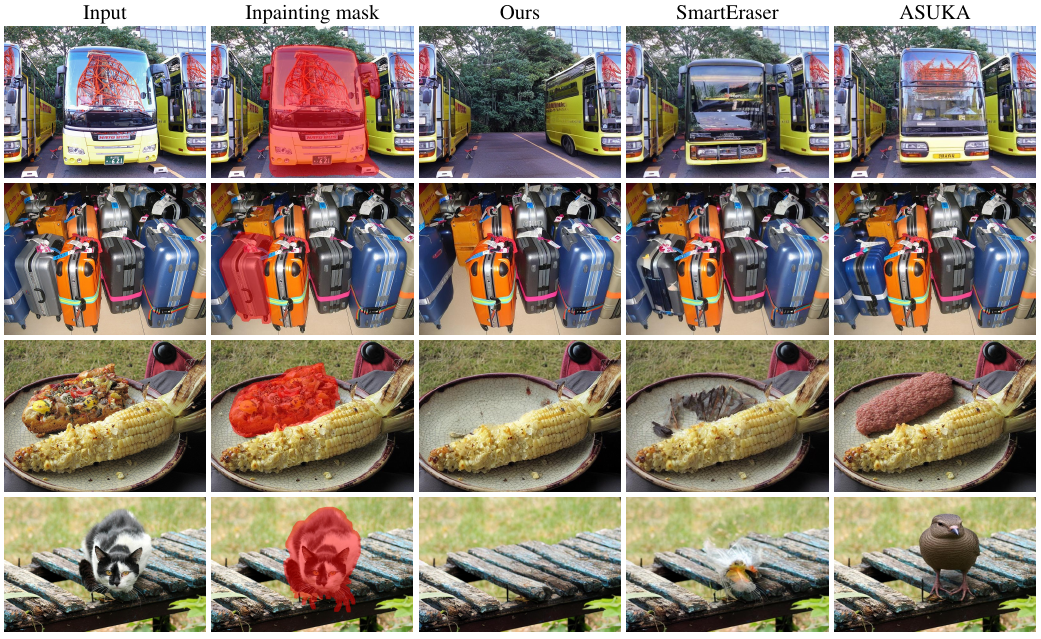}\\
    
    \vspace{-2mm}
    \captionof{figure}{\textbf{Comparison of object erasure.} Unlike the compared methods which tend to generate unwanted objects or artifacts within the masked regions, our method allows to cleanly erase the target object in a single pass, without introducing any unexpected content while maintaining the overall harmony and context consistency. }
    
    \label{fig:compare_top}
    
    \end{center}
    \bigbreak
]

\renewcommand{\thefootnote}{}
\footnotetext{$^*$Corresponding author (zhangq93@mail.sysu.edu.cn).}

\begin{abstract}

We present YOEO, an approach for object erasure. Unlike recent diffusion-based methods which struggle to erase target objects without generating unexpected content within the masked regions due to lack of sufficient paired training data and explicit constraint on content generation, our method allows to produce high-quality object erasure results free of unwanted objects or artifacts while faithfully preserving the overall context coherence to the surrounding content. We achieve this goal by training an object erasure diffusion model on unpaired data containing only large-scale real-world images, under the supervision of a sundries detector and a context coherence loss that are built upon an entity segmentation model. To enable more efficient training and inference, a diffusion distillation strategy is employed to train for a few-step erasure diffusion model. Extensive experiments show that our method outperforms the state-of-the-art object erasure methods. Code will be available at \href{https://zyxunh.github.io/YOEO-ProjectPage/}{https://zyxunh.github.io/YOEO-ProjectPage/}.

\end{abstract}    
\section{Introduction}
\label{sec:intro}

Recent advances in diffusion models~\cite{ddpm,diffusion_beat_gan,sd1.5,chen2024pixartalpha,podell2024sdxl,labs2025flux1kontextflowmatching} have greatly advanced image inpainting, enabling applications such as restoration, completion, and editing. Among these, object erasure has become an essential capability for visual editing and content refinement. Previously, most erasure methods relied on GAN-based frameworks~\cite{gan_first}, which often produced blurry textures and artifacts. In contrast, diffusion-based approaches can generate visually realistic results with rich details and textures, and have therefore become the mainstream solution.

However, diffusion-based methods often hallucinate by inserting unintended objects after removing the target ones, leading to contextually inconsistent results (see Figure~\ref{fig:compare_top}). On the other hand, recent closed-source multimodal models such as ChatGPT and Nano Banana, though are more powerful in object erasure, but involve large parameter counts and high computational overhead, hindering their practical deployment on edge devices. Hence, it is quite necessary to develop a dedicated object erasure model that not only enables superior erasure performance but also enjoys low inference latency and significantly fewer parameters.

Several specialized erasure methods try to mitigate hallucination issues by modifying attention probability maps~\cite{magiceraser,erase_diffusion}, generating synthetic data~\cite{jiang2025smarteraser}, 
introducing segmentation~\cite{zhu2025entityerasure} or geometric~\cite{,zhu2025georemover} priors. 
However, these methods still struggle in complex open-world scenes, often producing incoherent or unrealistic content. We identify two main causes of this limitation.
First, most diffusion-based erasure models are trained exclusively on synthetic paired data created by partially masking or inserting objects into clean images and using the original image as the ground truth. Such data fail to represent real-world erasure scenarios, where ground-truth erased results are unavailable.
Second, existing methods rely solely on Supervised Fine-Tuning (SFT), which trains diffusion models to reconstruct clean images from noisy inputs using pixel-level losses such as MSE or LPIPS. While effective for image generation, this procedure does not explicitly teach the model the task-specific goal of erasure, i.e., removing an object while preserving contextual consistency. Consequently, the model learns only to denoise rather than to erase coherently.

To address these limitations, we propose You Only Erase Once (YOEO), a framework capable of removing selected objects in a single pass while maintaining contextual and semantic coherence.
Our key insight is to incorporate pair-free, erasure-related supervision that guides the model to suppress unwanted content and enhance scene consistency.
Inspired by reward-based supervision~\cite{pick_reward,aligning_reward,raft_reward_rank,image_reward_nips,Human_preference_score_reward,ren2024byteedit}, we introduce a sundries detector that evaluates whether the generated image contains unwanted artifacts. Rather than relying on costly and inconsistent human annotations, we leverage a pre-trained entity segmentation model~\cite{entity_seg_iccv} as an automatic sundries detector~\cite{zhu2025entityerasure}.

However, applying this detector directly within a standard multi-step diffusion model is infeasible as the early denoising stages produce blurred images that obscure sundries detection~\cite{erase_diffusion}. To overcome this, we first distill a multi-step diffusion model into a few-step student model, enabling it to produce clear results at early denoising steps~\cite{Diff_instruct_distill,zhou2024score_distill}. This allows end-to-end optimization with sundries-aware supervision.
Additionally, to ensure semantic consistency, we measure the feature coherence between the completed and original regions using features extracted from a pretrained segmentation network. If the generated region aligns well with its surrounding context, its features should cluster around the same representation center.

In summary, our main contributions are as follows:
\begin{itemize}[leftmargin=2em]
\setlength\itemsep{0.5em}
\item We propose YOEO, a novel object erasure framework that allows to erase target objects in a single pass without introducing unwanted content while preserving semantic and contextual coherence.
\item We introduce a few-step distilled diffusion architecture that enables efficient training with unpaired real-world data for enhancing the erasure performance.
\item We design a sundries suppression loss to avoid generating unwanted content and an entity feature coherence loss to enforce consistency between erased regions and the surrounding context.
\end{itemize}
\begin{figure*}[!h]
        \centering
        \centering
        \adjustbox{scale=0.96,center}{\includegraphics[width=0.92\linewidth]{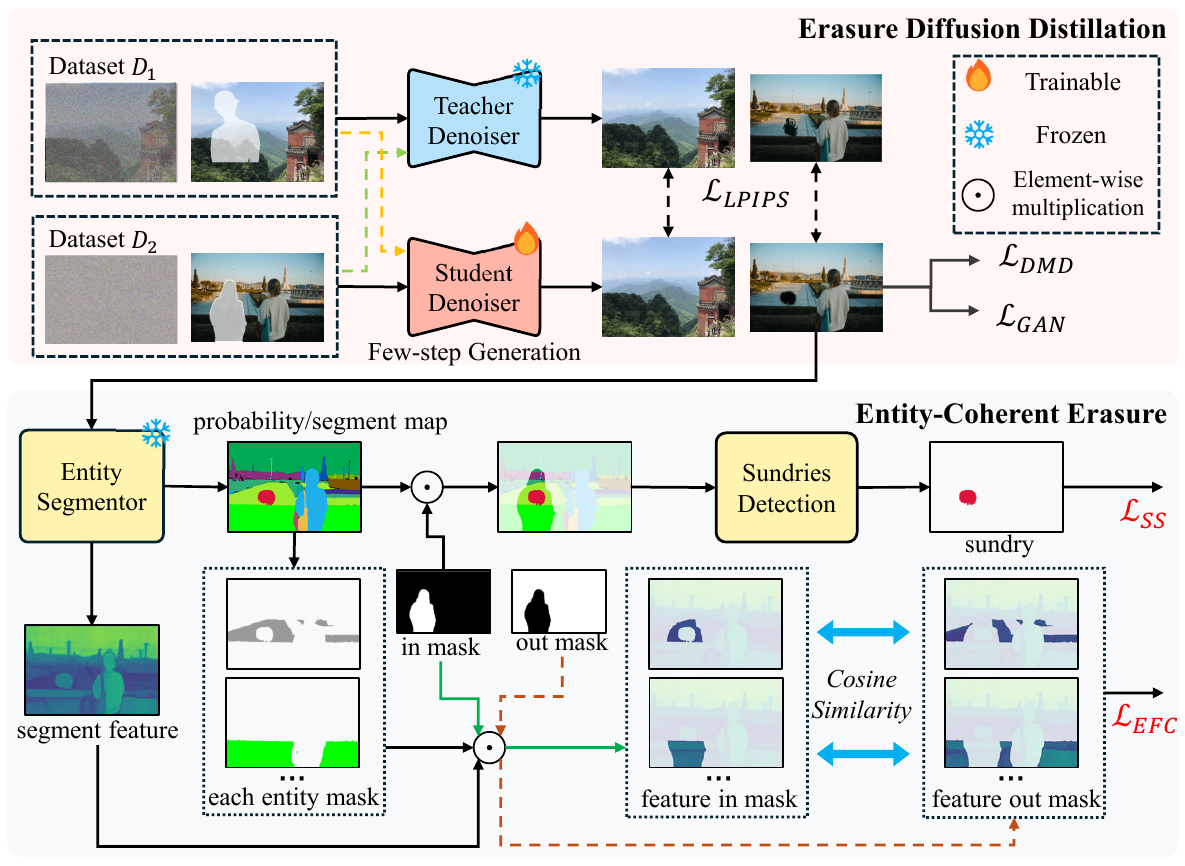}}

        \vspace{-3mm}
        \caption{\textbf{Overview of framework.} 
        It distills the student denoiser with datasets $\mathcal{D}_1$ and $\mathcal{D}_2$. The $\mathcal{D}_1$ is constructed by randomly occluding inpainting masks onto background regions, where the original image can be used as the corresponding ground truth. The $\mathcal{D}_2$ uses the objects in the image as inpainting masks, aiming to erase the selected objects. Erasure Diffusion Distillation is the basic distillation framework, introducing the distillation-related losses $\mathcal{L}_{LPIPS}$, $\mathcal{L}_{DMD}$, and $\mathcal{L}_{GAN}$. Entity-Coherent Erasure employs an entity segmentor to predict the entity segment of the erased image. The sundry entities filtered by Sundries Detection~(Fig.~\ref{fig:sundries}) serve as sundries suppression loss $\mathcal{L}_{SS}$ during training to suppress the generation of unwanted sundries. In addition, we compute the cosine similarity between the segment feature inside the generated region~(in mask) and those outside the region (out mask) as entity feature coherent loss $\mathcal{L}_{EFC}$, encouraging the model to generate context-consistent content.
        }
        \vspace{-1em}
        \label{fig:pipeline}
\end{figure*}

\section{Related Work}

\noindent \textbf{Image inpainting.} 
Early approaches~\cite{Incremental_transformer_gan_inpaint,yu2019free_gated_conv_gan_inpaint,cao2022learning_prior_gan_inpaint,contextual_attentioin_gan_inpaint,zhao2021large_gan_inpaint,dual_path_gan_inpaint,zits++_gan_inpaint} are mostly designed based on Generative Adversarial Networks (GANs)~\cite{gan_first}. While effective for small or structured missing areas, GAN-based methods often produce unrealistic textures and visible artifacts. Recently, large-scale text-to-image diffusion models~\cite{ddim,diffusion_beat_gan,sd1.5,podell2024sdxl,labs2025flux1kontextflowmatching} have opened a new path to perform image inpainting through iterative denoising~\cite{ddpm}, which are adopted to produce compelling inpainting results~\cite{chen2024anydoor,zhu2024logosticker,yu2025structure,powerpaint,Omnipaint,wang2025metashadow}. In contrast, object erasure aims to remove specified objects while seamlessly completing the missing regions without introducing hallucinated artifacts. Some recent methods work by incorporating prompt-based background guidance~\cite{ekin2024clipaway,powerpaint}, or geometric and amodal priors~\cite{zhu2025entityerasure,zhu2025georemover}.
ASUKA~\cite{asuka} combines masked autoencoders (MAE) with diffusion models to reduce hallucinated content, whereas SmartEraser~\cite{jiang2025smarteraser} synthesizes paired training data by pasting and uses the selected objects as prompts to explicitly indicate which objects should be removed. However, these methods still struggle to handle complex scenes and often produce results with unexpected content or contextual inconsistency.
\vspace{0.5em}
\noindent \textbf{Diffusion distillation.} As generating high-quality visual content with diffusion models requires many time-consuming denoising steps, various works focus on improving the computational efficiency of diffusion models. A solution is to distill a single-step student model to match the output distribution of a multi-step teacher model~\cite{Fast_Sampling_of_Diffusion_Model}, but the distilled model often yields blurry or over-smoothed results. To mitigate this, adversarial objectives have been integrated into the distillation process~\cite{xiao2021tackling_gan,sd_turbo,Ufogen_diffusion_gans,Semi_implicit_denoising_diffusion,YOSO}. Another line of work employs score distillation to align the distributions of teacher and student diffusion models~\cite{Diff_instruct_distill,dmd,zhou2024score_distill,TDM}. DMD2~\cite{DMD2} incorporates a GAN loss to correct the teacher’s score estimation errors, achieving enhanced realism.
To further improve efficiency and stability, Flash Diffusion~\cite{flash_diffusion} introduces a progressive noise-scheduling strategy that gradually increases the corruption level during training, demonstrating impressive performance across diverse tasks including inpainting, super-resolution, and face editing. TurboFill~\cite{turbofill_inpainting_insert} similarly adopts a three-stage adversarial training scheme for few-step diffusion inpainting and object insertion.
\vspace{-0.5em}
\section{Our Approach}

\vspace{-0.5em}

Given a pretrained erasure diffusion model $G_{init}$ serving as a teacher denoiser, we fine-tune our model into a few-step erasure model $G_{\theta}$ using a dataset containing paired samples $\mathcal{D}_1 = \{(X^i, Y^i, M^i)\}_{i=1}^{N_1}$ and unpaired samples $\mathcal{D}_2 = \{(X^j, M^j)\}_{j=1}^{N_2}$, where $X$ denotes the input image, $Y$ the target image, and ${M}$ the inpainting mask. To enable pair-free supervision for real-world erasure scenarios, we introduce two novel objectives: the sundries suppression (SS) loss and the entity feature coherence (EFC) loss. These losses complement the conventional distillation objectives, jointly guiding the model to remove specified objects cleanly while maintaining contextual and semantic consistency. The SS loss encourages the model to suppress unintended “sundry” generations, whereas the EFC loss enforces coherence between the erased region and its surrounding entities at the feature level. By integrating these objectives with distillation-based training, our approach enables efficient and precise object erasure within a few denoising steps. 
Below we elaborate on the details of our method.

\begin{figure}[t]
        \centering
        \centering
        \adjustbox{scale=1.07,center}{\includegraphics[width=0.94\linewidth]{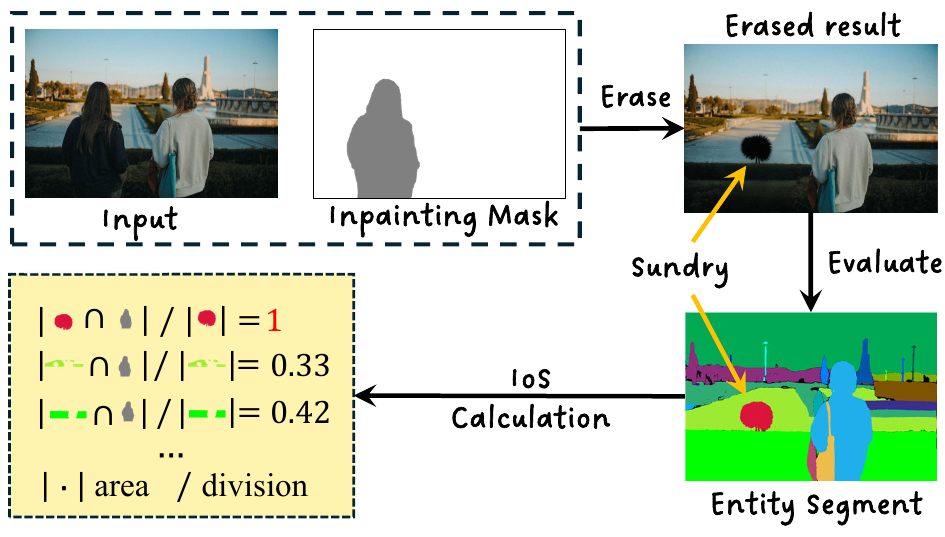}}
        \vspace{-7mm}
        \caption{\textbf{Illustration of our sundries detection.} To identify unintended objects (sundries) in the erased output, we first perform entity segmentation using a pretrained entity segmentation model. For each detected entity, we compute its IoS with respect to the inpainting mask. Entities whose IoS exceeds a threshold $\lambda$ are classified as newly generated sundries.}
        \label{fig:sundries}
        \vspace{-1em}
\end{figure}

\subsection{Dataset Construction}
Our training data is constructed to support both paired and pair-free supervision, enabling the model to learn effective object erasure from both synthetic and real-world scenarios.
We build the dataset upon 1.75M images from the Open Images dataset~\cite{open_image}, with text prompts set to empty. To incorporate fine-grained semantic annotations, we leverage the entity masks provided by EntityErasure~\cite{zhu2025entityerasure}. The final dataset consists of two complementary subsets: a synthetic paired dataset $\mathcal{D}_1$ and a real-world unpaired dataset $\mathcal{D}_2$.

\vspace{0.5em}
\noindent \textbf{Synthetic paired dataset.}\quad The paired subset is firstly used to facilitate supervised fine-tuning of the teacher diffusion model. Following the same strategy as EntityErasure~\cite{zhu2025entityerasure} and LaMa~\cite{lama}, we generate inpainting masks by randomly occluding partial regions of the background in the image. Each paired sample $(X^i, Y^i, M^i)$ is constructed as: $X^i = (1 - M^i) \odot Y^i$, where $Y^i$ is the ground-truth image and $M^i$ is the inpainting mask.
This process produces diverse and controllable occlusions while retaining pixel-level correspondence between inputs and ground-truth completions. 
In addition to supervising the training of the teacher model $G_{init}$, this data is also used to distill the student model, enabling stable distillation from the pretrained teacher model.

\vspace{0.5em}
\noindent \textbf{Real-World unpaired dataset}\quad 
To promote generalization to realistic object-erasure scenarios, we also construct an unpaired subset $\mathcal{D}_2 = \{(X^j, M^j) \mid M^j \in M^{entity}\}_{j=1}^{N_2}$, where each inpainting mask $M^j$ is sampled directly from entity masks $\mathbf{M}_{entity}$ in the original image.
Unlike the paired subset, the masked input $X^j$ has no corresponding ground-truth erased result. Instead, the student model learns to perform erasure using pair-free supervision provided by our proposed SS and EFC losses, which guide the network to remove the masked entities cleanly while preserving contextual consistency.

\subsection{Erasure Diffusion Distillation}

Our inpainting diffusion distillation framework is built upon Flash Diffusion~\cite{flash_diffusion}. The teacher diffusion model $G_{init}$ and the student diffusion model $G_{\theta}$ share the same network architecture, with the student initialized from the teacher’s parameters.  We first align the student’s single-step output with the teacher denoiser’s multi-step prediction using the following distillation loss:
\begin{align}
\mathcal{L}_{distill}&=\mathbb{E}\left[d(\text{ODE}(x_{t}, G_{init}, t, \mathbf{c}), G_{\theta}(x_{t},t, \mathbf{c}))\right],
\end{align}
where $\text{ODE}$ denotes the teacher’s multi-step prediction process, $t$ is a randomly sampled noise timestep, $
\mathbf{c} = (x_{m},m_{in})$ represents the condition composed of the masked latent $x_{m}$ and the inpainting mask $m_{in}$, $x_{t}$ is the noisy latent at timestep $t$, and $d(\cdot,\cdot)$ is a distance function, for which we adopt LPIPS~\cite{lpips} by default.

Beyond the distillation loss, we employ Distribution Matching Distillation (DMD) to ensure distributional consistency between the student and teacher denoisers. DMD achieves this by minimizing the Kullback–Leibler (KL) divergence between the real and generated distributions. The gradient of this objective is expressed as:
\begin{equation}
\nabla 
\mathcal{L}_{DMD}=\mathbb{E}\left[ -\left(s_{real}(y) - s_{fake}(y)\right)\nabla G_{\theta}(x_{t},t,c) \right],
\end{equation}
where $y$ denotes the latent generated by the student model, and $s_{real}$ and $s_{fake}$ are the score functions corresponding to the teacher and student distributions, respectively. 

We also use a GAN loss~\cite{gan_first} to improve the quality of images generated by the student diffusion model. It is applied directly in the latent space, with features extracted by the pre-trained teacher diffusion model serving as input to the discriminator. The GAN loss can be expressed as:
\begin{align}
\mathcal{L}_{GAN}&=0.5 \times \mathcal{L}_{G} + 0.5 \times \mathcal{L}_{D}, \nonumber \\
\mathcal{L}_{G}&=\mathbb{E}\left[ \left\|  D(G_{\theta}(x_{t},t,c)) - 1 \right\|^2 \right], \nonumber \\
\mathcal{L}_{D}&=\mathbb{E}\left[ \left\|  D(x_{0}) - 1 \right\|^2 + \left\|  D(G_{\theta}(x_{t},t,c)) \right\|^2\right],
\end{align}
where $\mathcal{L}_{G}$ is the generator~(student denoiser) loss, and $\mathcal{L}_{D}$ is the discriminator loss.

\subsection{Entity-Coherent Erasure}
If the student denoiser is distilled using only distillation losses, it will inevitably inherit the characteristics of the teacher denoiser. Consequently, the erased results may exhibit the same artifacts and are prone to generating sundries (unwanted residuals) in complex scenes. To address this issue, we aim for the distilled student erasure model to remove target objects without introducing undesired content. The completed regions should seamlessly extend the surrounding background or adjacent objects, while maintaining structural and semantic coherence with the scene.

\vspace{0.5em}
\noindent \textbf{Erasure-related supervision.} As shown in Figure~\ref{fig:pipeline}, we introduce two pair-free erasure-related supervision to enhance the student model during distillation: (1) sundries suppression~(SS) Loss which employs a sundries detector to identify and penalize the presence of unwanted content in the generated images, effectively reducing noise and undesired insertions; (2) entity feature coherence~(EFC) Loss that enforces consistency between the completed regions after erasure and their surrounding context in the original image, promoting spatial and semantic coherence in the restored scene. Together, these two losses guide the student model toward clean, contextually consistent object erasure results.

\vspace{0.5em}
\noindent \textbf{Sundries suppression Loss} is designed to suppress the generation of unwanted artifacts (sundries). In Figure~\ref{fig:sundries}, an object is considered a sundry if the Intersection over Self (IoS)~\cite{zhu2025entityerasure, entity_seg_iccv} between its overlap with the inpainting mask and its total area exceeds a threshold. Given the image $\hat{X}$ generated by our few-step denoiser, the pretrained Mask2Former~\cite{mask2former} segmentor $S_\theta$ predicts entity probabilities and segmentation masks:
\begin{equation}
\left( \mathbf{P},\mathbf{M} \right)=S_{\theta}(\hat{X}),
\end{equation}
where $\mathbf{P} = \{(p^i_{0}, p^i_{1})\}_{i=1}^{n}$ represents the negative ($p^i_{0}$) and positive ($p^i_{1}$) probabilities of $n$ entities. $\mathbf{M} \in \mathbb{R}^{n\times h \times w}$ denotes the predicted segmentation masks. Sundries indices are determined as:
\begin{equation}
\mathcal{I} = \{ i \mid  IoS(\mathbf{m}^i,\mathbf{m}_{in}) > \lambda, \: p^i_1 > \tau  \},
\end{equation}
where $\mathbf{m}^i$ is the $i_{th}$ predicted mask and $\mathbf{m}_{in}$ is the inpainting mask. The probability threshold $\tau$ is set to 0.2 and the IoS threshold $\lambda$ is set to 0.9. IoS is defined as:
\begin{equation}
IoS(\mathbf{m}^i,\mathbf{m}_{in})=\frac{|\mathbf{m}^i\cap\mathbf{m}_{in}|}{|\mathbf{m}^i|},
\end{equation}

To eliminate the identified sundries, we enforce the model to increase the negative (non-sundry) probability and simultaneously penalize pixel activations in predicted sundry masks. The sundries suppression loss $\mathcal{L}_{SS}$ is formulated as:
\begin{align}
\mathcal{L}_{SS} &= \sum_{i \in \mathcal{I}} \left[ - \alpha^{i} \log p^i_0  - \sum_{x,y}\frac{\log (1 - m^i_{x,y})}{h\times w} \right],
\end{align}
where $m^i_{x,y}$ denotes a pixel at position $(x, y)$ of the $i_{th}$ entity mask and $\alpha^{i}$ is a weighting coefficient, computed as the proportion of the image area occupied by the $i_{th}$ entity.

\vspace{0.5em}
\noindent \textbf{Entity feature coherence loss.}\quad 
Modern segmentation models generate masks using final-layer features, either via convolution layers~\cite{long2015fully} or matrix multiplication~\cite{mask2former}. Features corresponding to pixels of the same entity naturally form coherent clusters. Leveraging this, the EFC loss ensures that completed regions align semantically and structurally with the original entity.

Let $\mathbf{F}^{seg} \in \mathbb{R}^{h \times w \times c}$ denote the segmentation feature map from the last layer of Mask2Former, $\mathbf{m}_{in} \in \mathbb{R}^{h \times w}$ the inpainting mask, and $\mathbf{M}_{entity} \in \mathbb{R}^{n \times h \times w}$ the entity segmentation masks. The outpainting mask $\mathbf{m}_{out}$, corresponding to the visible (unmasked) region, is defined as the complement of $\mathbf{m}_{in}$. For each entity, we define its outpainting region as $R^i_{out}=\mathbf{m}_{out} \odot \mathbf{m}_{entity}^{i}$and compute the mean segmentation feature vector over this region to serve as the entity’s cluster center:
\begin{equation}
\mathbf{f}^{i} = \frac{1}{N_{pix}} \sum_{(x,y) \in R^i_{out}} \mathbf{F}^{seg}_{x,y},
\end{equation}
where $N_{pix}$ denotes the number of pixels in the outpainting region. 
\begin{figure*}[ht!]
    \centering
    \adjustbox{scale=1.01,center}{\includegraphics[width=1.0\linewidth]{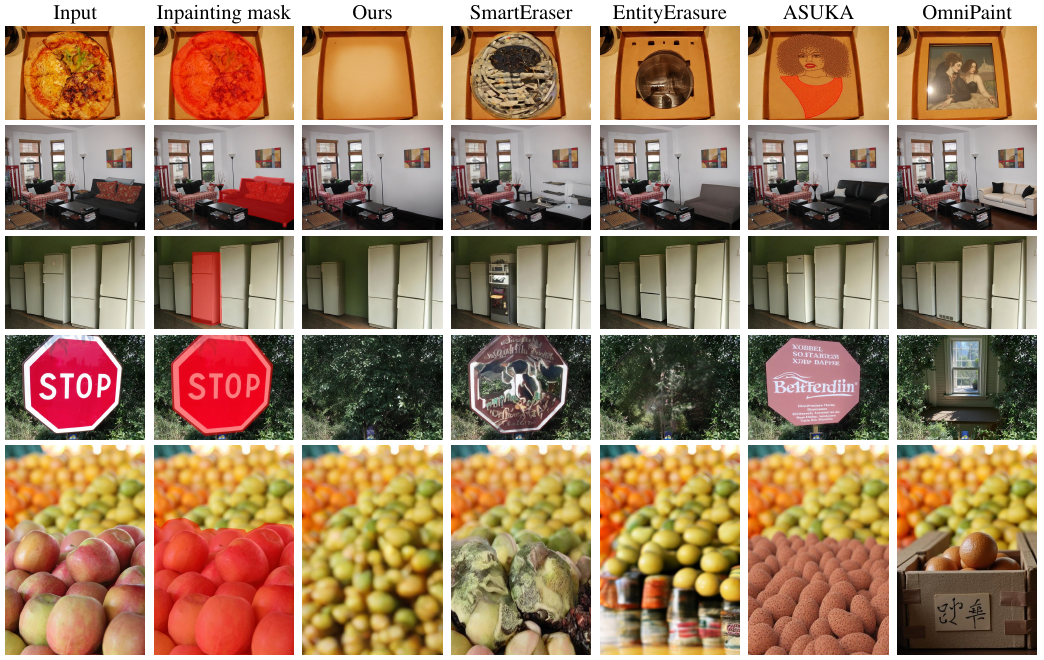}}
    \vspace{-0.6cm}
    \caption{\textbf{Qualitative comparison with state-of-the-art methods on the COCO dataset.}}
    \label{fig:compare_other}
    \vspace{-1em}
\end{figure*}
Our objective is to align the segmentation features within each inpainted region, defined as $R^i_{in}=\mathbf{m}_{in} \odot \mathbf{m}_{entity}^{i}$, with its corresponding cluster center using cosine similarity:
\begin{equation}
\mathcal{L}_{EFC} = - \sum_{i} \sum_{(x,y) \in R^i_{in}}
\frac{\mathbf{F}^{seg}_{x,y} \cdot \mathbf{f}^{i}}
{\|\mathbf{F}^{seg}_{x,y}\|\,\|\mathbf{f}^{i}\|}.
\end{equation}
This loss encourages dense features in the inpainted region to align with the original entity’s feature cluster, thereby promoting semantic and structural coherence between the generated and existing regions.

\subsection{Loss Function} 
During training, we combine teacher-student distillation with the erasure-related supervision to achieve entity-coherent completion. The overall loss function is expressed as:
\begin{equation}
\begin{split}
\mathcal{L}=&\lambda_{1}\mathcal{L}_{distill}+\lambda_{2}\mathcal{L}_{DMD}+\lambda_{3}\mathcal{L}_{GAN}\\&+\lambda_{SS}\mathcal{L}_{SS}+\lambda_{EFC}\mathcal{L}_{EFC},
\end{split}
\end{equation}
where we empirically set $\lambda_{1}=1$, $\lambda_{2}=0.7$, $\lambda_{3}=0.3$, $\lambda_{SS}=0.5$ and $\lambda_{EFC}=0.5$. $\mathcal{L}_{SS}$ and $\mathcal{L}_{EFC}$ are dynamically adjusted for stable training with the gradient normalization strategy used in VAE training~\cite{sd1.5}, and then multiplied by a scale factor $\lambda_{SS}$ and $\lambda_{EFC}$. 

\subsection{Implementation Details}

Our framework builds upon the pretrained Stable Diffusion Inpainting 1.5 model~\cite{sd1.5}. We employ the AdamW optimizer with a learning rate of $1\times10^{-5}$ and a weight decay of $0.01$. 
During inference, we use the LCM scheduler~\cite{lcm} with the number of steps set as 2.
All experiments are conducted on a single A100 GPU. Our training proceeds in the following three stages:

\vspace{0.5em}
\noindent \textbf{Stage 1: Teacher fine-tuning.}
We first fine-tune the teacher denoiser on the synthetic paired dataset using the standard diffusion training procedure. The batch size is set as $1000$, and the model is trained for $800$ iterations. This stage adapts the teacher model to the entity-erasure domain while maintaining the inpainting capability.

\vspace{0.5em}
\noindent \textbf{Stage 2: Student distillation.}
Next, we initialize the student denoiser by copying the weights of the fine-tuned teacher and distill it using paired data. The teacher uses the DPMSolver++ scheduler~\cite{dpm++} with 20 steps.
We randomly sample timesteps around ($249$, $499$, $749$, $999$), and apply a warm-up schedule that gradually increases the probability of selecting higher-noise timesteps.
Both models share the same optimizer settings. The batch size is set to $8$, and training is conducted for $15{,}000$ iterations.

\vspace{0.5em}
\noindent \textbf{Stage 3: Joint training with erasure-related supervision.}
Finally, we initialize the student denoiser with the distilled weights and train it jointly on both synthetic paired and real-world unpaired data. 
Since the unpaired data lacks ground truth, we feed pure noise at step $999$ during training and introduce extra pair-free losses.
The entity feature coherence loss is applied to both datasets, sundries suppression loss is applied only to the real-world data. We set the batch size to $7$ and train for $3{,}600$ iterations, while keeping all other hyperparameters the same as in the previous stage.

\begin{table*}[ht]
    \centering
        \caption{\textbf{Quantitative comparison with previous methods.} Markers \spc and \pct denote models based on stable-diffusion-1.5 and Flux.}
        \vspace{-2mm}
    \label{tab:compare_other}
    \begin{tabular}{@{}l|c|cccc|cccc|c} 
            \toprule
            & & \multicolumn{4}{c|}{EntitySeg~\cite{entity_seg_iccv}} & \multicolumn{4}{c|}{COCO~\cite{coco_dataset}} & \\
            Method & \#Params &FID\(\downarrow\) & MSN\(\downarrow\) & MARS\(\downarrow\) & CFD\(\downarrow\) & FID\(\downarrow\) & MSN\(\downarrow\) & MARS\(\downarrow\) & CFD\(\downarrow\) & Time(s)\(\downarrow\)\\
            \midrule
            MAT \cite{mat}& 62M & 50.0 & 0.433 & 0.044 & 0.457  &  25.2 & 1.88 & 0.133 & 0.773 & 0.07 \\
            \spc SDInpaint \cite{sd1.5} & 860M & 57.1 & 0.746 & 0.151&0.551 & 23.3 & 2.97 & 0.288& 0.857 & 1.75\\
            \spc CLIPAway \cite{ekin2024clipaway} & 860M & 59.0  & 0.679 & 0.130&0.504& 24.2  & 2.69 & 0.240& 0.809 & 3.57\\
            \spc PowerPaint \cite{powerpaint} &1746M & 56.9  & 0.554 & 0.115& 0.453 & 25.7  & 2.65 & 0.239& 0.792 & 4.09\\
            \spc SmartEraser \cite{jiang2025smarteraser}& 860M & 54.7  & 0.728 & 0.173& 0.577& 21.7  & 2.27 & 0.290 & 0.818 & 4.51 \\
            \spc EntityErasure \cite{zhu2025entityerasure}& 2608M& 58.0  & 0.122 & 0.037 & 0.363& 22.5  & 0.51 & 0.086& 0.584 & 2.38\\
            \pct ASUKA \cite{asuka} & 11902M & {\bf 47.6}  & 0.699 & 0.233& 0.565 & {\bf19.8}  & 1.49 & 0.318& 0.794  & 8.01\\
            \pct GeoRemover \cite{zhu2025georemover}& 12082M  & 52.0 & 0.600 & 0.146 & 0.508 & 23.8 & 1.37 & 0.116 & 0.684  &  68.4 \\
            \pct OmniPaint \cite{Omnipaint} &  11902M & 51.2  & 0.336 & 0.045 &0.407 & 22.1  & 1.47 & 0.111& 0.705  & 14.0\\
            \spc Ours & 860M & 58.8  & {\bf 0.049} & {\bf 0.005} & {\bf0.311} & 22.7  & {\bf 0.22} & {\bf 0.017}& {\bf 0.528} & 0.21\\ 
            \bottomrule
    \end{tabular}
    \vspace{-7pt}
\end{table*}

\section{Experiments}
\noindent \textbf{Datasets and metrics.} We evaluate the object-erasure performance on the COCO val2017 dataset~\cite{coco_dataset} and the Entity Segmentation test dataset~\cite{entity_seg_iccv}, both of which provide segmentation annotations. We exclude all “stuff” masks and use only “thing” (object) masks as inpainting masks to remove the corresponding target objects. The evaluation set includes 670 test samples from the Entity Segmentation dataset and 3,985 samples from COCO val2017. 
These are non-synthetic real-world data without ground-truth erased images, so we evaluate them using pair-free metrics. We use Mean Sundries Num (MSN) and Mean Area Ratio of Sundries (MARS)~\cite{zhu2025entityerasure} to evaluate whether the model generates additional unwanted objects, and CFD~\cite{Omnipaint} to evaluate context consistency. Compared with ReMove~\cite{remove_score}, CFD can additionally measure object hallucination. We also use FID~\cite{fid} to assess the overall image generation quality. 

\subsection{Comparison with State-of-the-Art Methods}

\noindent \textbf{Baselines.}\quad 
We compare our method with several state-of-the-art methods including MAT~\cite{mat}, SD-Inpainting~\cite{sd1.5}, PowerPaint~\cite{powerpaint}, CLIPAway~\cite{ekin2024clipaway}, EntityErasure~\cite{zhu2025entityerasure}, SmartEraser~\cite{jiang2025smarteraser}, ASUKA~\cite{asuka}, GeoRemover~\cite{zhu2025georemover} and OmniPaint~\cite{Omnipaint}. Among them, MAT is GAN-based, while all others are diffusion-based approaches.

\begin{table}[t]
    \centering
        \caption{\textbf{Ablation study.} $\mathcal{D}_2$ denotes the real-world unpaired dataset. ``$\checkmark$'' indicates that the component is enabled, while ``$\text{\ding{55}}$'' indicates it is disabled.} 
    \vspace{-0.2cm}
    \label{tab:ablation}
    \resizebox{\linewidth}{!}{
    \begin{tabular}{ccc|ccccc}
        \toprule
        EFC & SS & $\mathcal{D}_2$ & FID\(\downarrow\) & MSN\(\downarrow\) & MARS\(\downarrow\) & CFD\(\downarrow\)\\
        \midrule
        \xmark & \xmark & \xmark &  {\bf 57.2} & 0.330 & 0.0883 & 0.434  \\ 
         \xmark & \xmark & \cmark & 58.0 & 0.391 & 0.0873 & 0.461 \\
        \cmark & \cmark & \xmark &  58.0 & 0.193 & 0.0525 & 0.367  \\
        \cmark & \xmark & \cmark & 59.3 & 0.079 & 0.0174 &0.350  \\
        \xmark & \cmark & \cmark & 58.3 &  0.136 & 0.0168 & 0.354  \\
        \cmark & \cmark & \cmark & 58.8 & {\bf 0.049} & {\bf 0.0050} & {\bf 0.311} \\
        \bottomrule
    \end{tabular}
    }
    
\end{table}

\vspace{0.5em}
\noindent \textbf{Quantitative comparison.}\quad 
As shown in Table~\ref{tab:compare_other}, our method achieves substantial improvements over other methods 
on the sundries-related metrics MSN and MARS across both the EntitySeg and COCO datasets. This indicates that our approach effectively suppresses the generation of undesired objects. Furthermore, our method attains higher scores on the context consistency metric CFD, demonstrating its superior ability to preserve semantic coherence with the surrounding regions. Although our method is built on SD-1.5 with only 860M parameters, and thus has weaker inherent instruction-following ability compared with the 12B Flux-base model, it still achieves more context-consistent erasure while using far fewer parameters than Flux-based methods such as ASUKA, GeoRemover, and OmniPaint.

\vspace{0.5em}
\noindent \textbf{Qualitative comparison.}\quad 
Figure~\ref{fig:compare_other} presents visual comparisons with other state-of-the-art methods. As shown, our method avoids unwanted content and achieves more contextually coherent completions, particularly in complex scenes or those containing multiple interacting entities. Please see the supplementary material for more comparison results, and also the failure cases of our method.

\begin{figure}[t!]
    \centering
    \adjustbox{scale=1.01,center}{\includegraphics[width=1.0\linewidth]  {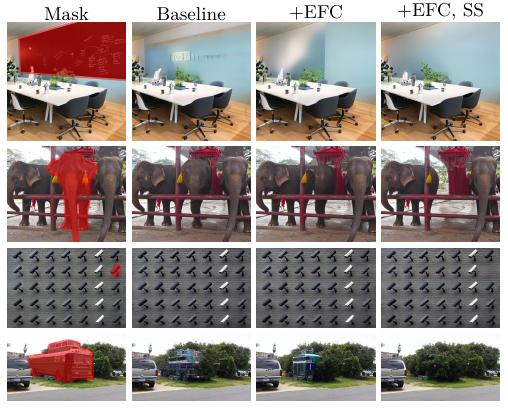}}\\
    \vspace{-4mm}
    \caption{\textbf{Qualitative ablation study of our method.} EFC and SS refer to the entity feature coherence loss and sundries Suppression loss, respectively. 
    The baseline indicates the student model distilled with $\mathcal{D}_2$, but without using the EFC or SS losses.}
    \label{fig:ablation}
    \vspace{-2mm}
\end{figure}

\subsection{Ablation Study}
Below we conduct ablation studies to evaluate the effectiveness of the components in our framework. Results are summarized in Figure~\ref{fig:ablation} and Table~\ref{tab:ablation}. 
Analysis on the sensitivity of loss-related hyper-parameters is given in the supplementary material.

\begin{figure}[t]
    \centering
    
    \adjustbox{scale=1.01,center}{\includegraphics[width=1.0\linewidth]{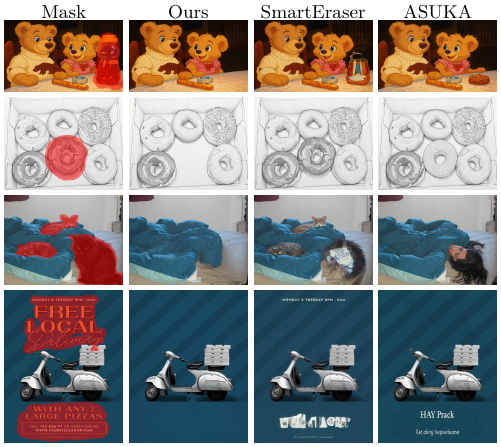}}\\
    \vspace{-2mm}
    \caption{\textbf{Comparison on diverse scenarios.} Our method works well for multi-object erasure as well as comics, sketches, and watercolor paintings.}
    \label{fig:more_scenes}
\end{figure}

\vspace{0.5em}
\noindent \textbf{Effect of entity feature coherence loss.}
We observe that the baseline model, trained solely with distillation-related losses, often introduces new unwanted sundries after erasure. Incorporating EFC loss encourages the model to generate content that remains semantically aligned with the surrounding context. This leads to more coherent and visually consistent completions, effectively reducing the occurrence of sundries. However, in complex scenes with dense object interactions, sundries may still exist.

\vspace{0.5em}
\noindent \textbf{Effect of sundries suppression loss.} 
Further integrating the sundries suppression loss, which explicitly leverages predicted sundry masks and their associated probabilities to penalize undesired content generation, results in a more substantial reduction of sundries. As shown in Figure~\ref{fig:ablation}, the qualitative results exhibit visibly cleaner erasure outputs. Quantitatively, Table~\ref{tab:ablation} confirms this trend, with notable decreases in sundries-related metrics (MSN and MARS), validating the effectiveness of the SS loss in suppressing unwanted object generation.

\vspace{0.5em}
\noindent \textbf{Effect of unpaired data $\mathcal{D}_2$.} 
Using SS and EFC losses only on the synthetic paired data 
$\mathcal{D}_1$ provides limited improvement, while additionally incorporating $\mathcal{D}_2$ for training is necessary to more effectively suppress sundry generation. As shown in Table~\ref{tab:ablation}, after adding $\mathcal{D}_2$, the sundries-related metrics~(MSN, MARS) and the context-coherent metric~(CFD) all decrease significantly, demonstrating the importance of training with unpaired data $\mathcal{D}_2$.

\begin{figure}[t!]
    \centering
    \adjustbox{scale=1.01,center}{\includegraphics[width=1.0\linewidth]{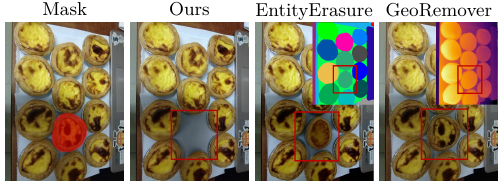}}\\
    \vspace{-2mm}
    \caption{
    \textbf{Comparison with condition-guided methods.} 
}
    \label{fig:compare_condition}
    \vspace{-0.3cm}
\end{figure}

\subsection{More Analysis}

\vspace{0.5em}

\noindent \textbf{Comparison with condition-guided methods.} Figure~\ref{fig:compare_condition} presents a comparison with condition-guided approaches. EntityErasure and GeoRemover respectively predict the post-erasure entity segmentation and depth map to guide subsequent image generation. However, these methods can fail when the initial condition prediction is inaccurate. In the egg tart example, both the segmentation predicted by EntityErasure and the depth predicted by GeoRemover produce unwanted tart contours, leading to incomplete erasure. 
In contrast, our method uses entity segmentation as an evaluator and integrates guidance directly into the generative training process, enabling effective erasure of the egg tart.

\begin{figure}[t]
    \centering
    \adjustbox{scale=1.0,center}{\includegraphics[width=1.0\linewidth]{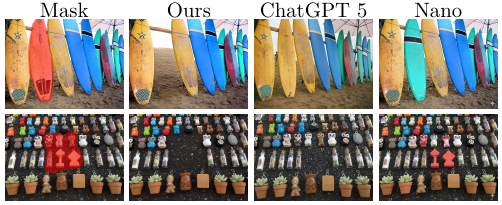}} \\
    \vspace{-2mm}
    \caption{\textbf{Comparison with powerful multimodal models.} }
    \label{fig:llm}
\end{figure}

\vspace{0.5em}
\noindent \textbf{Comparison with powerful multimodal models.} Figure~\ref{fig:llm} compares our method with ChatGPT-5 and Nano Banana on object erasure. As shown, our method produces cleaner and context-consistent erasure results, without introducing unexpected content, while the compared large multimodel models generate extra unwanted objects, manifesting the effectiveness of our method.

\vspace{0.5em}
\noindent \textbf{Generalization across diverse scenarios.} To assess the robustness of our method beyond natural images, we evaluate on domains including animations, artistic portraits, sketches, and posters. As shown in Figure~\ref{fig:more_scenes}, our method effectively removes target objects and synthesizes visually coherent completions,
demonstrating strong generalization capability across diverse domains.
\vspace{1em}
\vspace{0.1cm}
\section{Conclusion}
\vspace{0.1cm}
We have presented a novel diffusion-based object erasure framework that allows to erase target objects cleanly while completing the erased regions with contextually coherent content. Our method enhances erasure performance by incorporating erasure-related supervision during the distillation stage and effectively leveraging large-scale unpaired data without ground-truth erasure counterparts. Extensive experiments and ablation studies demonstrate that our method significantly outperforms current state-of-the-art methods and generalizes well to various image domains.

\noindent \textbf{Acknowledgement.} This work was supported by the National Natural Science Foundation of China~(62471499), the Guangdong Basic and Applied Basic Research Foundation~(2023A1515030002).

{
    \small
    \bibliographystyle{ieeenat_fullname}
    \bibliography{main}
}

\end{document}